\documentclass[final,5p,times,twocolumn]{elsarticle}

\usepackage{hyperref}
\usepackage{lineno}
\usepackage{comment}
\usepackage{amsmath}
\usepackage{float}
\usepackage{graphicx}
\usepackage{svg}
\usepackage{pdfpages} 
\modulolinenumbers[5]
\usepackage{pgfplots}
\usepackage{amssymb}
\usepackage{multirow}
\usepackage{gensymb}
\usepackage{multicol}
\usepackage{makecell}
\usepackage{siunitx}
\usepackage{tikz}
\usepackage{subfig}
\usepackage{caption}
\usepackage{array}
\usepackage[ruled,vlined]{algorithm2e}
\usetikzlibrary{matrix}
\DeclareMathOperator*{\argmax}{argmax}

\journal{arXiv}

\begin{document}
\begin{frontmatter}

\title{ \huge Deep multi-stations weather forecasting: explainable recurrent convolutional neural networks\\
}
\author{Ismail Alaoui Abdellaoui}
\author{Siamak Mehrkanoon\corref{cor1}}

\cortext[cor1]{Corresponding author}

\address{Department of Data Science and Knowledge Engineering, Maastricht University, The Netherlands}


\begin{abstract}
Deep learning applied to weather forecasting has started gaining popularity because of the progress achieved by data-driven models. The present paper compares two different deep learning architectures to perform weather prediction on daily data gathered from 18 cities across Europe and spanned over a period of 15 years. We propose the Deep Attention Unistream Multistream (DAUM) networks that investigate different types of input representations (i.e. tensorial unistream vs. multistream ) as well as the incorporation of the attention mechanism. In particular, we show that adding a self-attention block within the models increases the overall forecasting performance. Furthermore, visualization techniques such as occlusion analysis and score maximization are used to give an additional insight on the most important features and cities for predicting a particular target feature of target cities.
\end{abstract}

\begin{keyword}
Weather data \sep deep learning \sep convolutional neural network \sep attention mechanism \sep explainability
\end{keyword}
\end{frontmatter}
\section{Introduction}
Weather forecasting is of major importance as it affects the daily activities of fundamental fields such as agriculture, transportation and international commerce among others. 
The ability to forecast the precipitation rates, the risk of flood or the likelihood of a hurricane can potentially lead to saving of lives and to the well-being of humans. Moreover, the change of the climate on earth has led to increasing research and world-wide efforts to halt the environmental and ecological consequences \cite{o2017ipcc}.

Traditional approaches of weather forecasting rely on priors like the thermodynamic properties of the atmosphere \cite{holtslag1990high, niziol1995winter, campbell2005weather}, statistical distribution of the data \cite{glahn1985statistical}, or ensemble learning that incorporates multiple models with different initial conditions \cite{gneiting2005weather}. This family of models belongs to the ``Numerical Weather Prediction" (NWP) methodologies \cite{lorenc1986analysis} and usually rely on the processing power of supercomputers, thus being resource heavy \cite{bauer2015quiet}. In addition of the high computational cost, it has been shown that \textit{a priori} information about the data that constitutes the initial state is the source of errors in weather prediction \cite{tolstykh2005some}.

While traditional NWP methods aim at extracting useful dynamics from a model or to transfer information between models, the purpose of recent data-driven approaches is to simulate an entire system to predict its future state \cite{scher2018toward}. Machine learning data-driven based models have already been successfully applied in various domains such as healthcare, dynamical systems, biomedical signal analysis, neuroscience among others \cite{mehrkanoon2012approximate,mehrkanoon2015learning,mehrkanoon2014parameter,mehrkanoon2019deep,mehrkanoon2018deep, abdellaoui2020deep,breiman2001random,webb2018deep,mehrkanoon2019cross}. The recent advances of machine learning models has increased the capability to automatically learn the underlying nonlinear complex patterns of weather dynamics \cite{mehrkanoon2019deep2,trebing2020smaat,trebing2020wind}. In particular, the combination of convolutional neural networks (CNNs) and long short-term memory (LSTM) networks proved to be a successful deep learning approach for climate modeling and weather forecasting \cite{chen2019hybrid, fu2019multi}.

This paper presents three contributions. The first one is an investigation of the unistream and multistream approaches as input representation for the neural networks. The second contribution aims at enriching the proposed networks with a self-attention mechanism. Finally, the third contribution addresses the explainability of the networks by using modern visualization techniques to determine the features and cities that contribute the most to the output predictions of a particular city or group of cities. It is of utter importance to gain interpretability from the data driven models given that weather prediction is the basis of many human real-life decisions. This paper is organized as follows. A brief review of the existing machine learning methodologies for weather forecasting is given in Section \ref{sec:related_work}. A formal definition of the Conv-LSTM layer and the visualization techniques used are presented in Section \ref{sec:preliminaries}. Our proposed models are introduced in Section \ref{sec:proposed_models}. Furthermore, the dataset used is introduced in Section \ref{sec:data_desc}. The experimental results are reported in section \ref{sec:results}. Finally, a discussion followed by the conclusion are drawn in sections \ref{sec:discussion} and \ref{sec:conclusion}, respectively.

\section{Related Work}\label{sec:related_work}
Multiple approaches have been recently proposed to tackle weather forecasting using deep machine learning models. The author in \cite{mehrkanoon2019deep2} introduced the convolutional neural networks to learn the underlying spatio-temporal patterns of weather data. This work used hourly past data from cities in the Netherlands, Belgium and Denmark to predict the temperature and wind speed of multiple cities. It has been shown that the convolutional operations benefit from a tensorial representation in order to improve the prediction capability.

In another work, a feed forward neural network has been used to investigate the volume of data needed as well as its recency to yield accurate weather predictions \cite{booz2019deep}. 
In terms of data volume, it has been shown that more data consistently leads to better predictions. The impact of the data recency remains unknown as there was no significant impact on the predictions when tuning the recency of the data.

In \cite{zhou2019forecasting}, the authors used deep learning to predict weather phenomena related to the heating of the air (i.e. heavy rains and thunderstorms among others), more commonly known as severe convective weather (SCW). A deep CNN has been utilized and proved to yield superior results compared to traditional machine learning models such as support vector machines or random forests. Another model which used stacked ConvLSTM layers, \textit{DeepRain}, was compared to linear regression models and reduced the RMSE by a large margin. This model used past radar data with a 6-min time resolution over a period of two years \cite{kim2017deeprain}. 

Similarly to the previous work, the authors in \cite{sonderby2020metnet} used a multi-input network with past radar data to perform precipitation forecasting. However, a different approach was used since the regression problem was transformed into a multi-class classification of possible precipitation ranges. The model also made use of axial self-attention \cite{ho2019axial} as spatial aggregator. This model was able to outperform the system used in the National Oceanic and Atmospheric Administration (NOAA). In \cite{dueben2018challenges}, the different challenges of using deep learning for weather forecasting are addressed. In particular, it has been stated that while neural networks can be useful for short-term predictions, the need for domain knowledge is essential when tackling forecasting of longer term ranges.

\section{Preliminaries}\label{sec:preliminaries}
\subsection{Self-Attention}
The self-attention mechanism was first introduced by Vaswani et. al \cite{vaswani2017attention} to capture dependencies within sequence of words. It relies on the dot product operation to assess the similarity of each word with respect to all the other words of a sequence. The query $Q$, key $K$, and value $V$ matrices are computed through the sequence of inputs $I \in \mathbb{R}^{S \times E}$ where $S$ is the sequence length and $E$ is the embedding dimension of each input feature:
\begin{equation}
    \quad Q=IW_{q}, \quad K=IW_{k}, \quad and \quad V=IW_{v},
\end{equation}
where $W_q$, $W_k$, and $W_v$ are learnable weights through a linear function. The attention matrix, also called the head output, is then computed through the softmax of a scaled dot product as follows:
\begin{equation}\label{eq:att}
    Attention(Q,K,V) = softmax(\frac{QK^T}{\sqrt{d_k}})V,
\end{equation}
where $d_k$ is the dimension of the key vector $K \in \mathbb{R}^{1 \times d_k}$.

\subsection{Conv-LSTM} \label{ssec:conv_lstm}
In this section we give an overview of the ConvLSTM layer that is used in the proposed models. 

It is based on the LSTM cell and was introduced in \cite{xingjian2015convolutional} to address the issue of capturing the spatial structure of the data. In this model, the input gate $i_t$, the forget gate $f_t$, the output gate $o_t$, the hidden state $h_{t-1}$, the candidate cell state $\hat{C}_t$, the current cell state $C_t$ and the input $x_t$ are all 3D tensors. The first dimension of each tensor is the sequence length while the two last dimensions represent the rows and columns. This model has first been used on weather data for precipitation nowcasting, outperforming other models based on the LSTM only.

\subsection{Activation maximization} \label{ssec:activ_max}
Activation maximization is a visualization technique that looks for patterns that maximize a particular activation function inside a neural network \cite{mahendran2016visualizing}. Following the taxonomy of interpretability methods presented in \cite{molnar2020interpretable}, activation maximization is a post hoc method since it aims at understanding the model after the training. This methodology focuses on finding a new input that maximizes the activation of a neuron:
\begin{equation} \label{eq:act_max}
    I^* = \argmax_{I}h_{l,z}(I),
\end{equation}
where $I$ is the input data of the network, $h$ is the activation function used in the neuron $z$ of the layer index $l$. For our case, we want to find the input data that contributes the most to minimizing the error between the model prediction and ground truth data. Since in our study weather element forecasting is reduced to a regression problem, here we define $h$, a custom objective function, as the inverse of the mean squared error (MSE):
\begin{equation}\label{eq:objective}
    h = \frac{1}{\frac{1}{n}\sum_{i=1}^{n} (y_i-\hat{y}_i)^2},
\end{equation}
where $y_i$ and $\hat{y}_i$ are the true measured data and the model prediction of a particular weather feature for the i\textsuperscript{th} target city, respectively. Here, $n$ denotes the number of target cities. The pseudocode of maximizing the $h$ score and getting the score map $I^*$ is provided in Algorithm \ref{alg:act_max}.

\begin{algorithm}[]
\SetAlgoLined
\KwIn{The number of iterations $s$\\
\Indp \Indp Pretrained model $m_p$\\
Sample input $I$\\
Input ranges $I_{min}$ and $I_{max}$\\
Learning rate $\eta$}
\KwOut{New input $I^*$}
  \For{the number of iterations $s$}{
    Perform a forward pass of $m_p$ on $I$ to get a prediction $\hat{y}$.\\
    Use eq. (\ref{eq:objective}) to obtain the score $h$.\\
    Apply $L_2$ normalization on the obtained score.\\
    Compute the gradient $dI$ of the normalized score with respect to input $I$.\\
    Update the input using: $I \leftarrow I + \eta \; dI$.  \\
    }
Obtain $I^*$ by clipping $I$ based on $I_{min}$ and $I_{max}$.
\caption{Score Maximization}
\label{alg:act_max}
\end{algorithm}

\subsection{Occlusion Analysis}\label{ssec:occlusion_analysis}
The occlusion analysis is a simplistic, yet effective way to determine which features contribute the most to a minimal error between the actual and prediction data. In this paper, we are concerned with two types of occlusion analysis: a spatial and a temporal occlusion. While the spatial occlusion analysis focuses on the important cities and weather features, the temporal analysis aims at determining the most important lags. The spatial occlusion analysis can be used to either focus on the cities only, or on the weather features only, or on a group of cities and features. However, all of these approaches rely on the same principle, which is to compute the percentage change between a reference MSE, obtained from the prediction of an unmasked data sample and its corresponding ground truth target data, and a new MSE, computed from the prediction of a masked data sample and the same ground truth label. We perform this percentage change each time the mask is slided to a new location of the input. Focusing on either the cities or the weather features on one hand, or a group of cities and features on the other hand, will determine the shape of the mask (i.e. a vector or a matrix, respectively). We present in Algorithm \ref{alg:occ_ana} the specific pseudocode of the occlusion analysis, when using a square matrix of size $p$ as a mask over the input dataset $\mathcal{X}$, and for a particular target city.
\begin{algorithm}[!h]
\SetAlgoLined
\KwIn{Input dataset $\mathcal{X}=\{x_i\}_{i=1}^{k}$\\ \vspace{0.05in}
\Indp \Indp Target dataset $\mathcal{O}=\{o_i\}_{i=1}^{k}$ \\
Pretrained model $m_p$\\
Target city index $c$\\
Mask size $p$\\
Number of horizontal slidings $s_h$\\  
Number of vertical slidings $s_v$}  
\KwOut{Occlusion map $M_o$}
  \For{the number of data samples}{
    Perform a prediction of the sample $x_i$ using $m_p$.\\
    Compute MSE\textsubscript i between the real target data $o_i$ and the model prediction for the $c$\textsuperscript{th} city. \\
    \For{the number of horizontal slidings $s_h$}
    {
        \For{the number of vertical slidings $s_v$}
        {
        Mask the sample $x_i$ by the patch to get the masked sample $\tilde{x}_{i}$.\\
        Make a new prediction using masked sample $\tilde{x}_{i}$.\\
        Compute the $\widetilde{MSE}$\textsubscript{i} between $o_i$ and the recent model prediction for the $c$\textsuperscript{th} city.\\
        Calculate the percentage change $\Delta$ between MSE\textsubscript i and $\widetilde{MSE}$\textsubscript{i}.\\
        Store $\Delta$ in a tensor $\mathcal{M}$ for this particular patch location.\\
        Displace the mask vertically by the distance $p$ over the sample $x_i$.
        }
        Displace the mask horizontally by the distance $p$ over the sample $x_i$.
    }
    
    }
    For each mask location, compute an average of the stored $\Delta$ over all data samples from $\mathcal{M}$ to get the occlusion map $M_o$.
\caption{Occlusion analysis}
\label{alg:occ_ana}
\end{algorithm}

\section{Proposed Models}\label{sec:proposed_models}
The two proposed models aim at studying the impact of the input representation for the task of weather forecasting. To this end, these models use two types of input representation: a unistream tensorial representation and a multistream representation.

\subsection{Unistream model}\label{sec:shared_convlstm_model}
The input of the model is a tensor $\mathcal{T} \in \mathbb{R}^{L \times F \times C}$ where $L$ is the number of lags used, $F$ is the number of weather features, and $C$ the number of cities. This tensor is the input of a  ConvLSTM layer. As previously seen in section \ref{ssec:conv_lstm}, the ConvLSTM layer essentially processes this tensor input while taking the number of lags $L$ as the sequence length. This layer is then followed by batch normalization and a flatten operation since the output of the ConvLSTM is tensorial. We then use two fully connected layers with a $ReLu$ activation function before the output layer.

\subsection{Multistream model}
This architecture uses a multistream approach. Each input stream uses a tensor $\mathcal{U} \in \mathbb{R}^{V \times F \times C}$ where $V$ is the number of lags used in each tensor. Since the two models, i.e. Unistream and Multistream, use the same number of lags, then $V$ evenly divides the total number of lags $L$ in each sample. Two ConvLSTM layers are used in each stream to capture the spatial and temporal features. The output of each stream is then concatenated on the axis of the channels. Similarly to the previous model, we use batch normalization and layer flattening. A dense layer with a $ReLu$ activation function is used before the final output layer. Some hyperparameters like the kernel size or the number of fully connected nodes have been adapted to have a comparable number of paramaters with respect to the previous model.

\subsection{Attention enriched models}
The two models presented above have also been augmented with a self-attention mechanism. 
More specifically, one layer encoder block introduced in \cite{vaswani2017attention} has been incorporated. 

In the tensorial input model, it has been added right after the ConvLSTM layer. For the multistream approach, the attention block is inserted after the merging. For both approaches, a reshaping is necessary before the attention encoder block since it is designed for matrices. It should also be noted that we only use one head attention.
\newline

\begin{figure}[!t]
    \centering
    \includegraphics[scale=0.2]{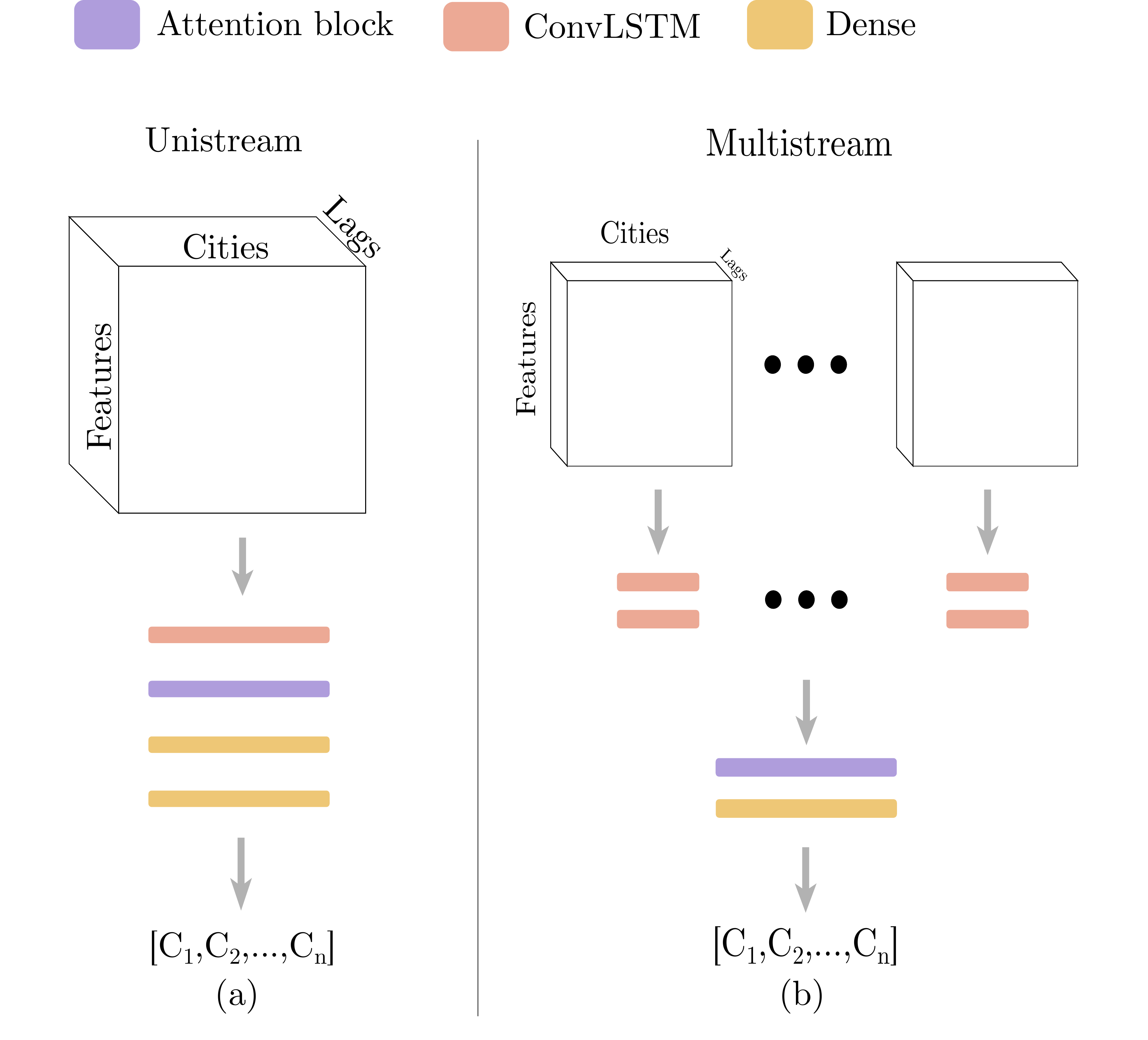}
    \caption{Schemas of (a) the Unistream and (b) Multistream models.}
    \label{fig:all_models}
\end{figure}

Fig. \ref{fig:all_models} shows the schema of the proposed models. The output of these models is a vector $o$ of length $n$, i.e. the number of target cities. Each value in this vector represents the same target feature for all the target cities. The hyperparameters of these models were selected so that they all have a comparable number of learnable parameters.

\section{Data Description}\label{sec:data_desc}
\begin{figure}[!t]
\centering
\includegraphics[height=3.3in,width=\columnwidth]{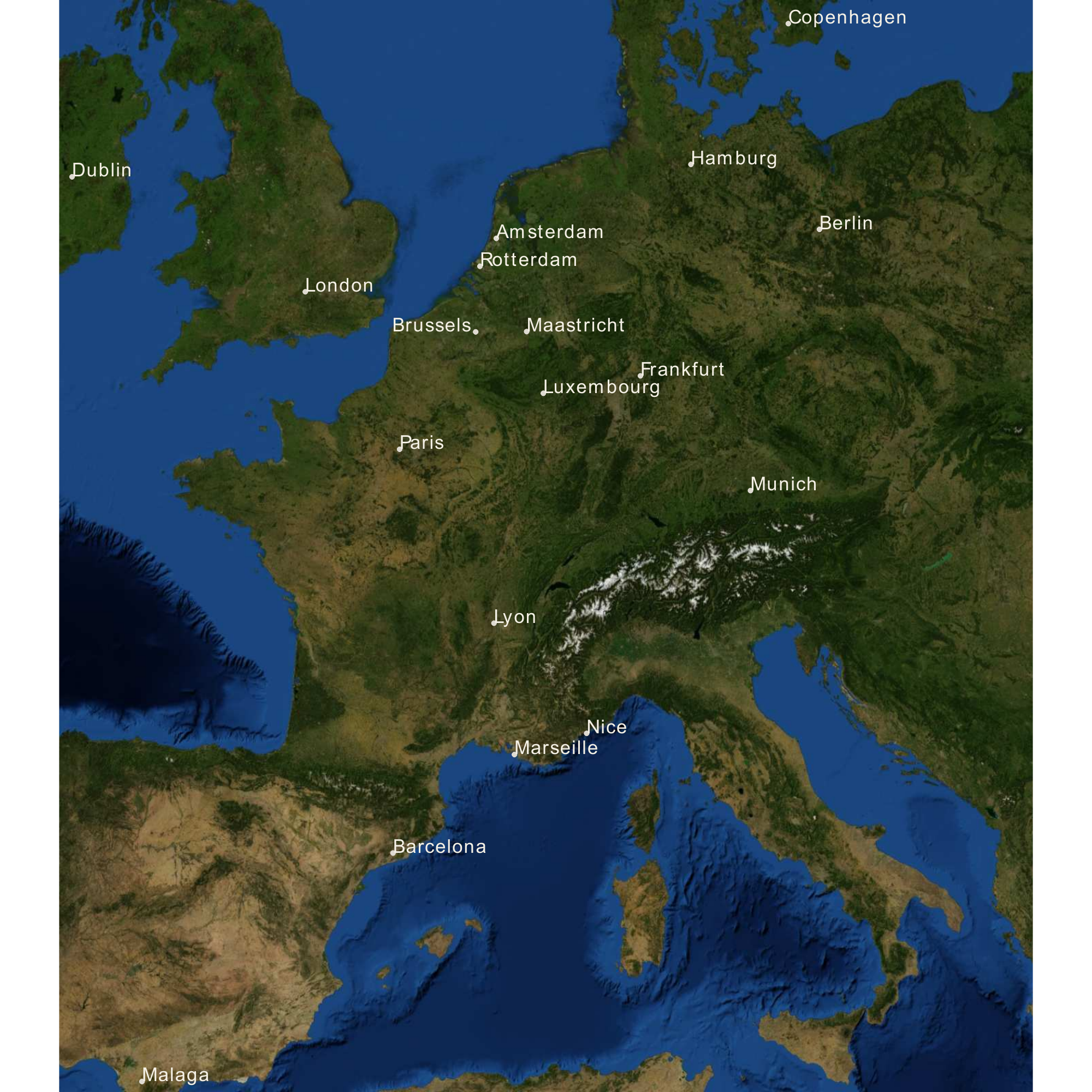}
\caption{Map showing the 18 cities used in the dataset.}
 \label{fig:map_high_res}
\end{figure}

The dataset used has been collected from Weather Underground and includes 18 cities across Europe and 18 weather features, for a period of 15 years from May 2005 to April 2020. The data is made publicly available \footnote{\url{https://github.com/IsmailAlaouiAbdellaoui/weather-forecasting-explanable-recurrent-convolutional-NN}}. Its time resolution is daily and weather features include for instance the temperature, wind speed, condition and sea level pressure among others. Table \ref{tab:features_explanation} presents the list of all features used. At each time step $t$, a data sample is represented by a matrix $M_t\in \mathbb{R}^{F \times C}$, where $F$ is the number of features and $C$ is the number of cities. Therefore the whole dataset is a tensor $\mathcal{D} \in \mathbb{R}^{L \times F \times C}$, where $L$ is the total number of days used. Fig. \ref{fig:map_high_res} shows a map of the different cities that are contained in the dataset. 
\begin{table}[!t]
\begin{center}
\caption{Features used in the dataset.}
\resizebox{\columnwidth}{!}{%
\centering
 \begin{tabular}{c >{\centering\arraybackslash}m{0.5\textwidth}}
 \Xhline{3\arrayrulewidth} 
 \multirow{2}{*}{\textbf{Feature name}} & \multirow{2}{*}{\textbf{Remarks}} \\
 & \\
 \Xhline{3\arrayrulewidth} 
 Highest temperature (\degree F) & -\\
 Lowest temperature (\degree F) & -\\
 Average temperature (\degree F)  & -\\
 Dew point (\degree F) & - \\
 Highest dew point (\degree F) & - \\
 Lowest dew point (\degree F) & - \\
 Average dew point (\degree F) & - \\
 Maximum wind speed (mph) & -\\
 Visiblity (mi) & Discrete value expressed in miles to measure the distance at which an object can be clearly distinguished\\
 Sea level pressure (Hg) & Measured in inch of mercury\\
 Observed temperature (\degree F) & Temperature in Fahrenheit observed at 10 am\\
 Observed dew point (\degree F) & Dew point in Fahrenheit observed at 10 am\\
 Humidity (\%) & -\\
 Wind direction & Discrete values indicating 16 possible directions of the wind \\
 Wind speed (mph) & -\\
 Wind gust (mph) & -\\
 Pressure (in) & -\\
 Condition & 21 possible discrete values that describe the overall weather state (cloudy, rainy, fog ...)\\
\Xhline{3\arrayrulewidth}  
\label{tab:features_explanation}
\end{tabular}
}
\end{center}
\end{table}

\section{Experimental Results}\label{sec:results}
\subsection{Data Preprocessing}\label{sec:Preprocessing}
The weather data is first scaled by means of equation \ref{eq:minmax}. In this way, for each feature and city, we take the values corresponding to every date and scale them down between 0 and 1.
\begin{equation}\label{eq:minmax}
    x_{scaled} = \frac{x - min(c_{ij})}{max(c_{ij}) - min(c_{ij})},
    i \in [1,F], j \in [1,C],
\end{equation}
where $c_{ij} \in \mathbb{R}^{L}$ refers to a column vector and $i$ and $j$ are the i\textsuperscript{th} feature of the j\textsuperscript{th} city.
\subsection{Experimental setup}
For all the experiments, the following six cities have been used as target cities: Paris, Luxembourg, London, Brussels, Frankfurt and Rotterdam. We also selected two target features: the wind speed, in miles per hour, and the average temperature of the day, in degree Fahrenheit. It should be noted that for each training instance, the output vector corresponds to only one type of feature, for all the target cities. Moreover, we performed experiments for 2, 4 and 6 days ahead, using 10 lags.
All the experiments used $90\%$ of the data for training and validation, while the remaining $10\%$ was used for testing. Adam method \cite{kingma2014adam} is used to optimize the mean square error (MSE) with a learning rate of $1e^{-4}$ and a batch size of $16$ for all of the proposed models.

\subsection{Results} \label{ssec:results}
The obtained mean squared error (MSE) of the proposed two models for wind speed as well as average temperature prediction for six target cities over 2, 4, and 6 days ahead are tabulated in Table \ref{tab:results_windspeed_mse} and Table \ref{tab:results_avgtemp_mse} respectively. The results of the two models with incorporated attention mechanism is also tabulated in Table \ref{tab:results_windspeed_mse} and Table \ref{tab:results_avgtemp_mse}. For every city and days ahead, the MSE of the best model is underlined. It should be noted that the reported MSEs are calculated after descaling the models prediction.
\begin{table}[!t]
    \centering
    \caption{The MSE comparison of the four models, for 2, 4, and 6 days ahead \textbf{wind speed} prediction.}
    \label{tab:results_windspeed_mse}
    \resizebox{\columnwidth}{!}{%
    \begin{tabular}{l l c c c c}\Xhline{3\arrayrulewidth}
    \multirow{2}{*}{\textbf{Days ahead}}&
    \multirow{2}{*}{\textbf{City}}&
    \multirow{2}{*}{\textbf{Unistream}}& \multirow{2}{*}{\textbf{Att-Unistream}}& \multirow{2}{*}{\textbf{Multistream}}& \multirow{2}{*}{{\textbf{Att-Multistream}}}\\
    & & & & & \\\Xhline{3\arrayrulewidth}
      2 & Luxembourg&28.85&\underline{22.96}&26.94&23.05\\
 & Rotterdam&48.04&\underline{38.22}&44.85&38.38\\
 & Frankfurt&41.11&\underline{22.70}&38.38&32.84\\
 & Brussels&36.78&\underline{29.26}&34.34&29.38\\
 & London&30.75&\underline{24.46}&28.71&24.56\\
 & Paris&25.25&\underline{20.09}&23.58&20.16\\ \hline
    4 & Luxembourg&32.27&25.82&29.85&\underline{25.35}\\
 & Rotterdam&53.73&42.98&49.69&\underline{42.20}\\
 & Frankfurt&45.97&36.78&42.52&\underline{36.11}\\
 & Brussels&41.14&32.91&38.05&\underline{32.31}\\
 & London&34.39&27.51&31.80&\underline{27.01}\\
 & Paris&28.24&22.59&26.12&\underline{22.18}\\ \hline
    6 & Luxembourg&38.22&26.34&30.63&\underline{25.36}\\
 & Rotterdam&63.64&43.87&51.00&\underline{42.23}\\
 & Frankfurt&54.45&37.53&43.64&\underline{36.14}\\
 & Brussels&48.72&33.58&39.05&\underline{32.33}\\
 & London&40.73&28.07&32.64&\underline{27.03}\\
 & Paris&33.45&23.06&26.81&\underline{22.20}\\ \hline
        \Xhline{3\arrayrulewidth}
    \end{tabular}
    }
\end{table}
From Tables \ref{tab:results_windspeed_mse} and \ref{tab:results_avgtemp_mse}, one can observe that often the models with attention are the most successful ones. Indeed, for both the prediction of the temperature and the wind speed, the models with attention always have an edge over their corresponding models without attention.

\begin{figure*}[!t]
\centering
\subfloat[]{\includegraphics[width=\columnwidth]{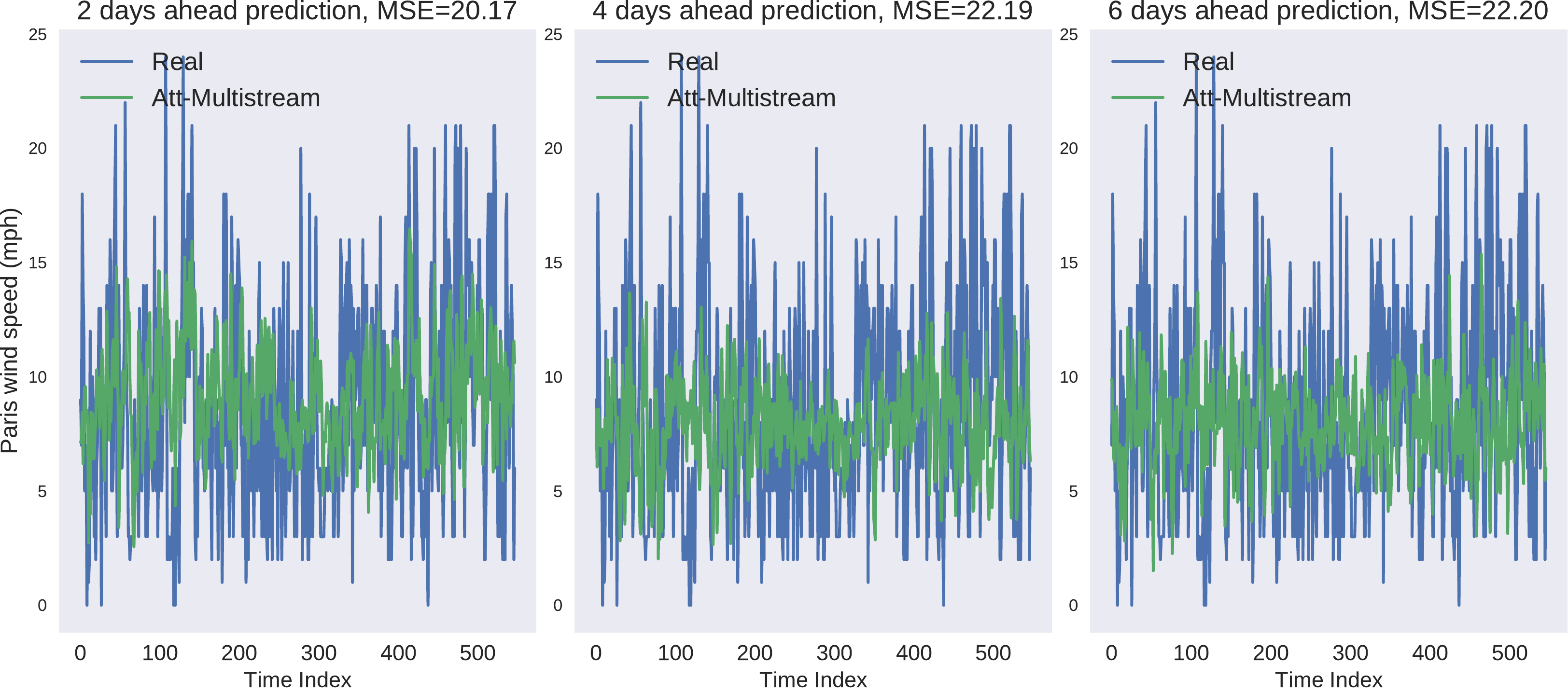}}
\subfloat[]{\includegraphics[width=\columnwidth]{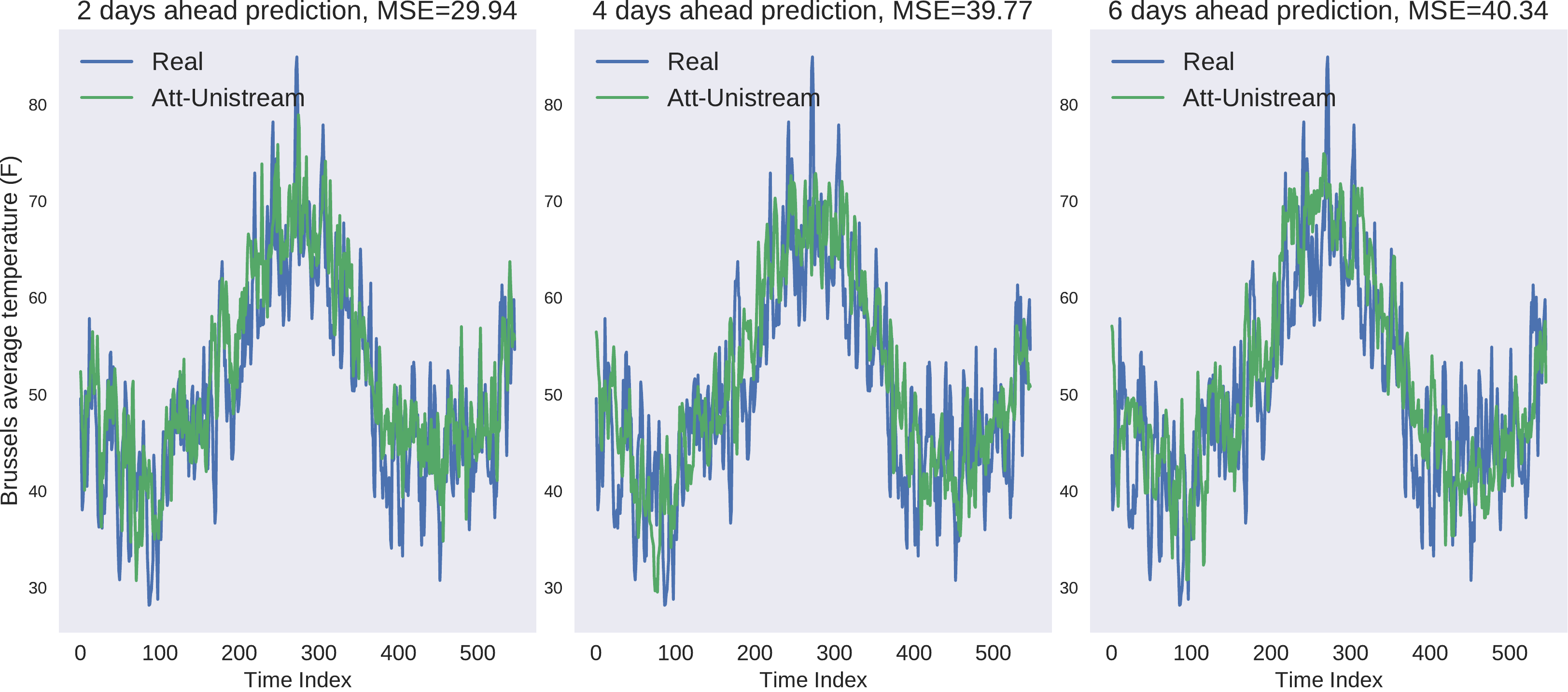}}

\caption{Actual vs. prediction of the average temperature (a) and wind speed (b) using the two proposed models.}
\label{fig:actual_vs_pred}
\end{figure*}

If we consider the models without the addition of attention, the one that uses a multistream approach is the most dominant one for both weather features, consistently outperforming the unistream model. Concerning the models with attention, if we compare them over all the results, there is no clear winner. 
However, if we perform the same analysis within each weather feature, a different pattern emerges. Indeed, the unistream model is the best one for predicting the temperature. It should also be noted that the unistream model is better than the other one for short time horizons (e.g. 2 days ahead). The city for which wind speed is easier to predict is Paris, while the average temperature of Brussels is the one yielding the minimum MSE error among all the target cities. Fig. \ref{fig:actual_vs_pred} shows the result of real data versus its prediction for 2, 4 and 6 days ahead using the Att-Multistream model and for the cities of Paris and Brussels.

\begin{table}[!t]
    \centering
    \caption{The MSE comparison of the four models, for 2, 4, and 6 days ahead \textbf{average temperature} prediction.}
    \label{tab:results_avgtemp_mse}
    \resizebox{\columnwidth}{!}{%
    \begin{tabular}{l l c c c c }\Xhline{3\arrayrulewidth}
    \multirow{2}{*}{\textbf{Days ahead}}&
    \multirow{2}{*}{\textbf{City}}&
    \multirow{2}{*}{\textbf{Unistream}}& \multirow{2}{*}{\textbf{Att-Unistream}}& \multirow{2}{*}{\textbf{Multistream}}& \multirow{2}{*}{{\textbf{Att-Multistream}}}\\
    & & & & & \\\Xhline{3\arrayrulewidth}
      2  & Luxembourg&58.40&\underline{40.83}&46.35&47.46\\
 & Rotterdam&52.89&\underline{37.14}&41.98&43.23\\
 & Frankfurt&53.73&\underline{37.68}&42.65&43.89\\
 & Brussels&42.87&\underline{30.15}&34.02&35.18\\
 & London&44.80&\underline{31.48}&35.56&36.69\\
 & Paris&53.15&\underline{37.27}&42.19&43.41\\ \hline
    4 & Luxembourg&67.71&53.88&59.41&\underline{42.16}\\
 & Rotterdam&61.32&48.91&53.81&\underline{38.19}\\
 & Frankfurt&62.29&49.67&54.66&\underline{38.80}\\
 & Brussels&49.70&39.70&43.61&\underline{30.97}\\
 & London&51.94&41.46&45.58&\underline{32.36}\\
 & Paris&61.62&49.13&54.07&\underline{38.29}\\ \hline
    6 & Luxembourg&75.87&\underline{54.91}&65.96&55.76\\
 & Rotterdam&68.72&\underline{49.84}&59.74&50.55\\
 & Frankfurt&69.80&\underline{50.60}&60.68&51.35\\
 & Brussels&55.69&\underline{40.43}&48.41&40.99\\
 & London&58.20&\underline{42.23}&50.60&42.84\\
 & Paris&69.05&\underline{50.06}&60.03&50.80\\ \hline
        \Xhline{3\arrayrulewidth}
    \end{tabular}
    }
\end{table}

Concerning the training time, we observed that it keeps growing as we add the attention mechanism and use the multistream approach. Interestingly enough, the multistream architecture that incorporates attention takes more time to train despite less trainable parameters. We should also note that the attention mechanism has more impact on the training time of the unistream architecture.

\section{Discussion} \label{sec:discussion}

An effective way to understand which input features and cities affect the outputs is using the techniques explained in sections \ref{ssec:activ_max} and \ref{ssec:occlusion_analysis}. In this section, we present the results of spatial and temporal occlusion analysis and score maximization techniques for the two proposed models. In addition, for this analysis the models have been trained to predict six days ahead.

\subsubsection{Occlusion analysis}
In order to determine which features contribute the most to a minimal error between the actual data and the prediction for a particular city, we first determine the MSE between the prediction of a sample and the actual data, which is used as a reference MSE. We then use a mask vector $m_f \in \mathbb{R}^{1 \times F}$ which is used in a sliding fashion across all the feature rows of the input data. We make an inference everytime $m_f$ masks a row, compute the corresponding MSE and finally obtain the percentage change between this MSE and the reference MSE. We repeat the same process along all row features to obtain all the percentage change for that particular data sample. The masked row feature that leads to the biggest MSE increase corresponds to the most important feature. Moreover, we repeat the same computations using multiple data samples and we average the percentage changes for each feature row. The same process applies to the rest of the other cities in order to obtain their corresponding important features. On the other hand, to determine the most important cities, we use the same algorithm, but with a mask vector $m_c \in \mathbb{R}^{C \times 1}$ that is slided across all the column cities of the input data.

\begin{figure*}[!htbp]
\centering
\subfloat[]{{\includegraphics[scale=0.067]{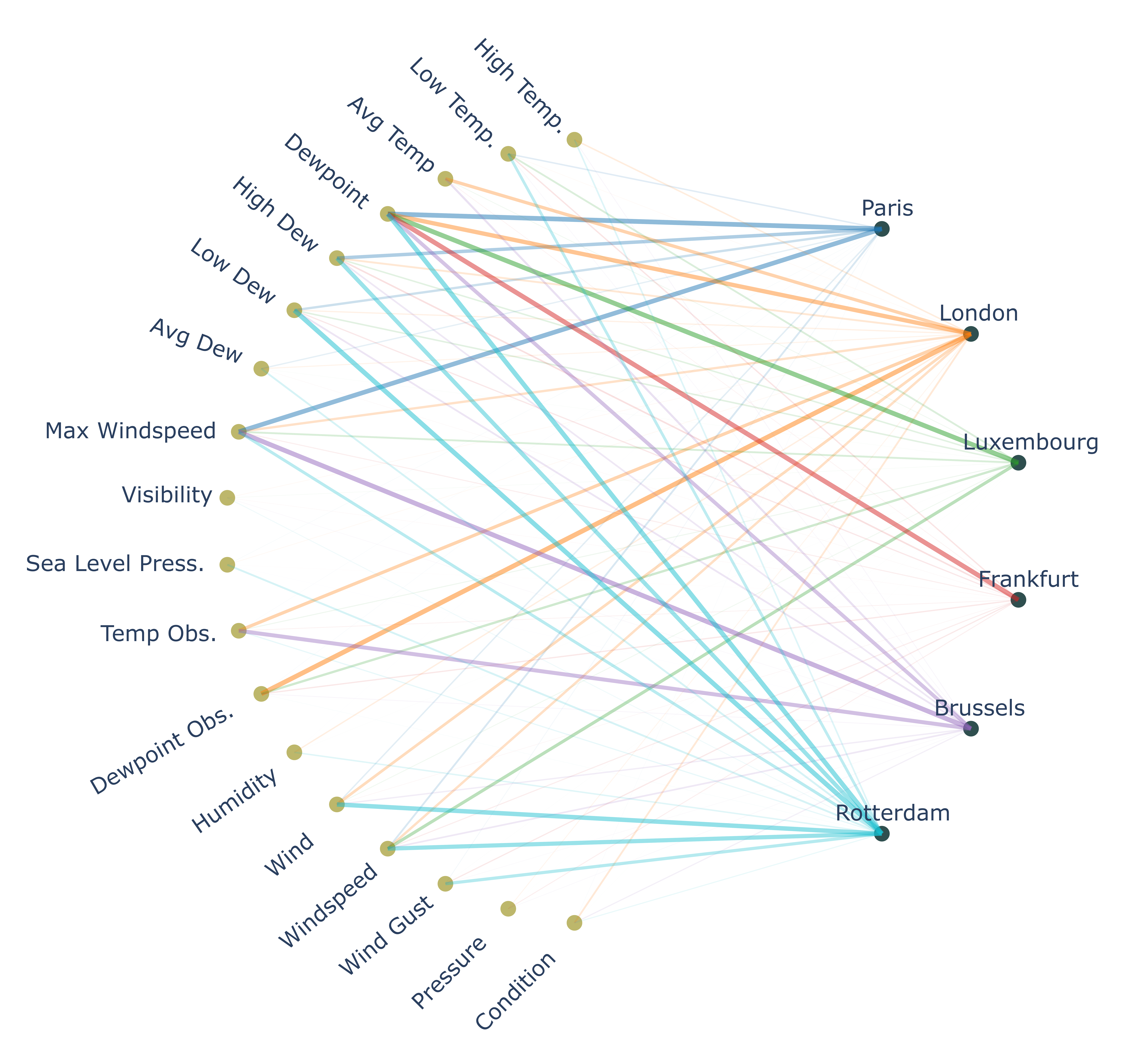}}}
\subfloat[]{{\includegraphics[scale=0.067]{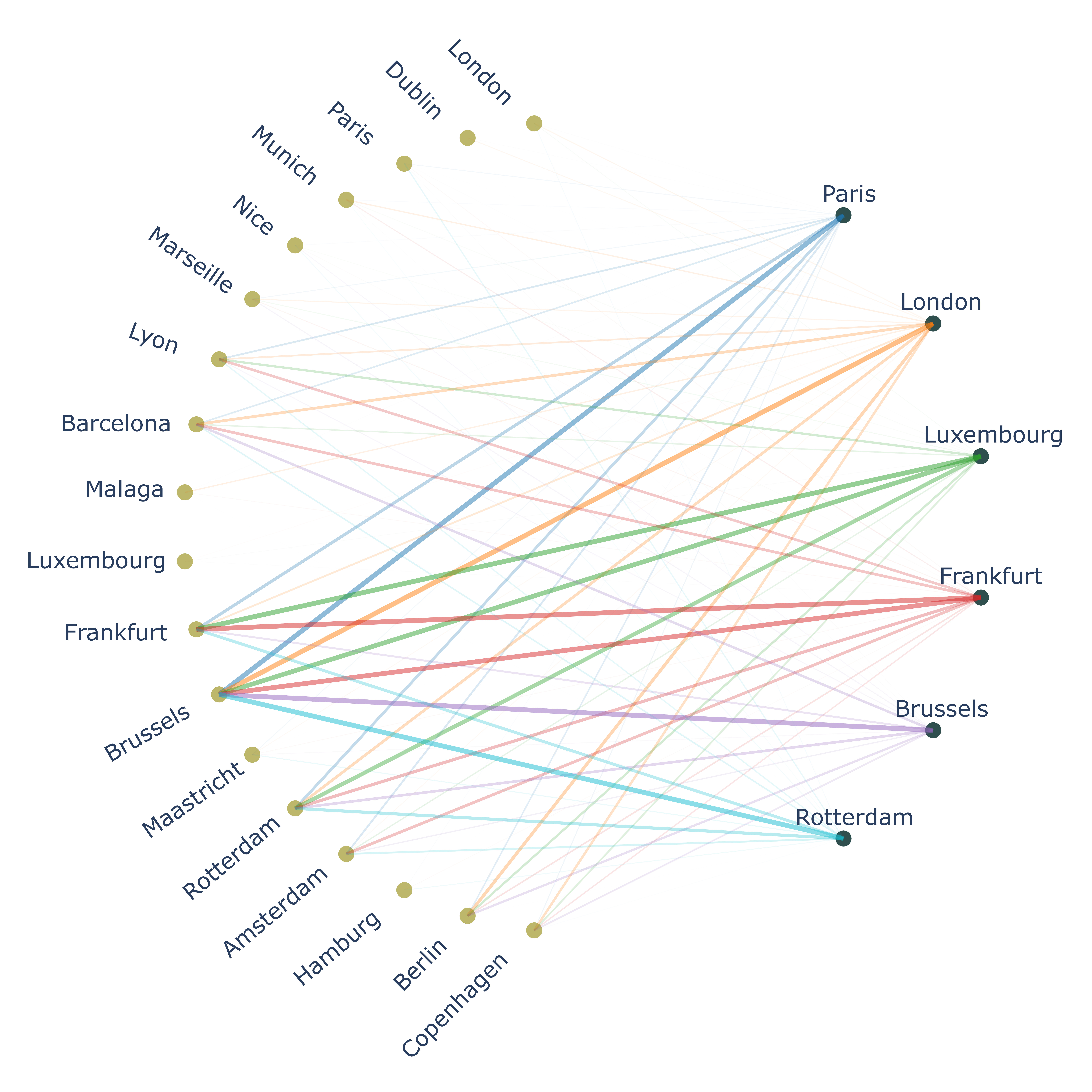}}}
\caption{Occlusion analysis visualization of the Att-Unistream model showing the most relevant weather features (a) and cities (b) for each target city. The model was trained to predict the 6 days ahead average temperature.}
\label{fig:occ_analysis_m1}
\end{figure*}

Fig. \ref{fig:occ_analysis_m1} shows the most important features and cities of the Att-Unistream model after performing the occlusion analysis. This model was trained on the 6 target cities described above with the average temperature as target feature. Fig. \ref{fig:occ_analysis_m1} (a) shows the most important features for each target city while Fig. \ref{fig:occ_analysis_m1} (b) presents the most relevant cities for each target city. Fig. \ref{fig:occ_analysis_m1} (a) highlights the importance of the dew point on the temperature since the plot shows an agreement around this weather feature. This feature makes sense since the dew point is the temperature to which air must be cooled to transform into water vapor. Concerning the cities, Brussels plays a major role into the predictions. This finding is reasonable since we can observe from Fig. \ref{fig:map_high_res} that Brussels is the centroid of the city cluster that makes up the target cities. Indeed, these models are multioutput, and affecting the city of Brussels would affect the prediction of all the target cities.

\begin{figure}[!htbp]
\centering
\includegraphics[scale=0.065]{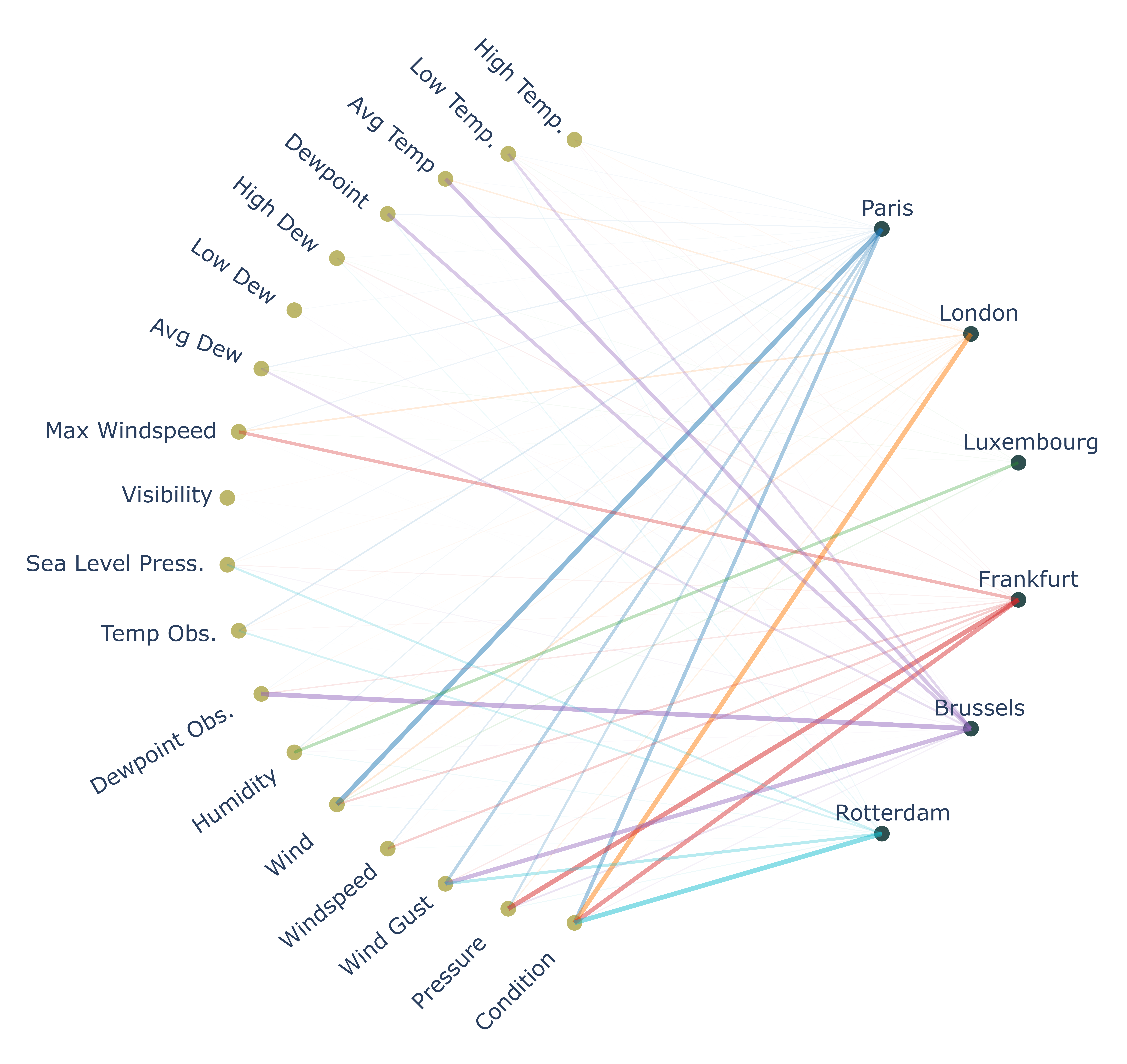}
\caption{Occlusion analysis visualization of the Att-Multistream model  showing the most relevant weather features for each target city. The model was trained to predict the 6 days ahead wind speed.}
\label{fig:occ_analysis_m3}
\end{figure}

Fig. \ref{fig:occ_analysis_m3} presents the same information as in Fig. \ref{fig:occ_analysis_m1}, but for the Att-Multistream model, with the wind speed as target feature. Interestingly enough, the condition and to a lesser extent the wind gust and pressure play some role in these predictions.

\begin{figure}[!t]
\centering
\subfloat[]{{\includegraphics[scale=0.18]{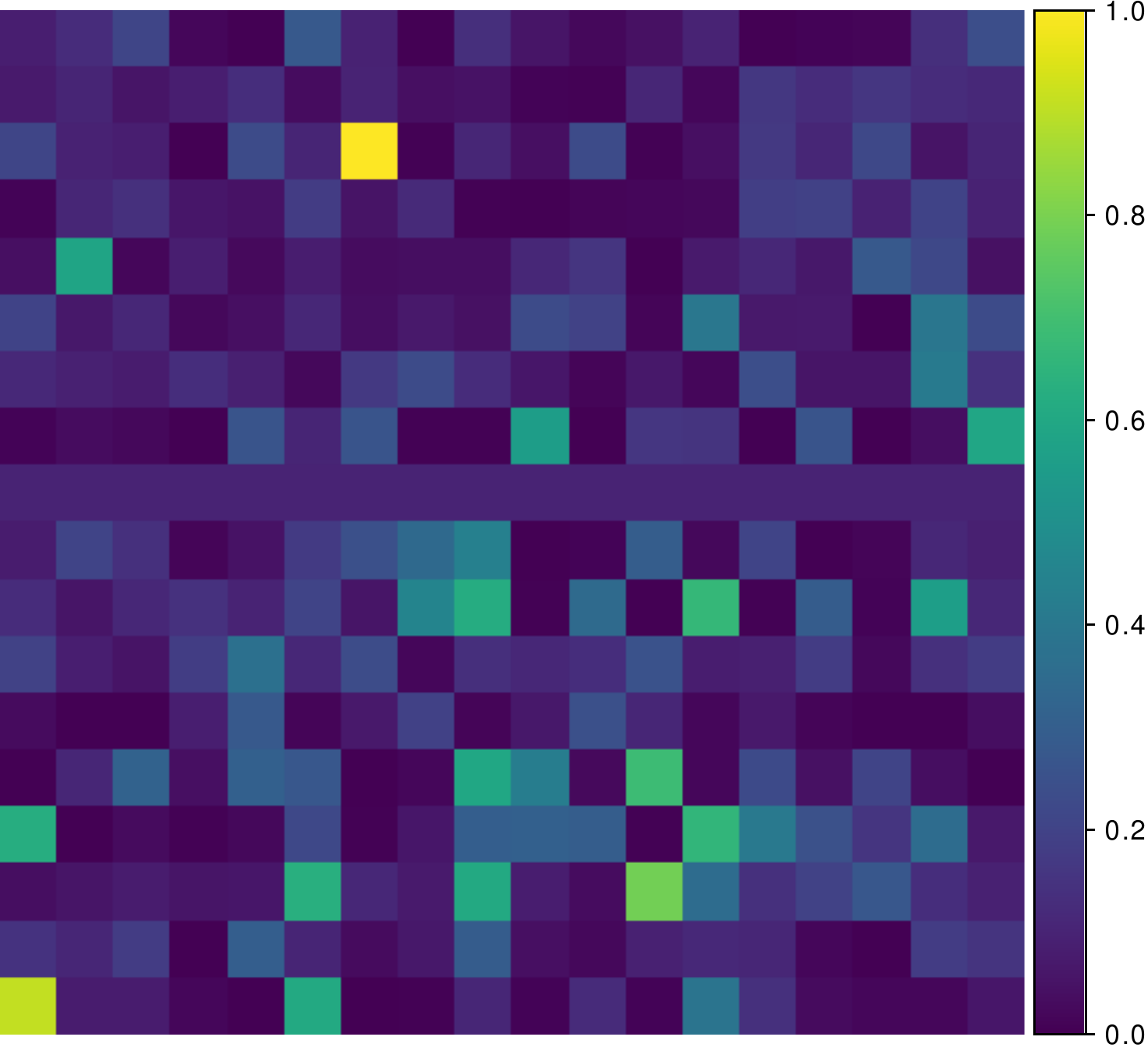}}}
\subfloat[]{{\includegraphics[scale=0.18]{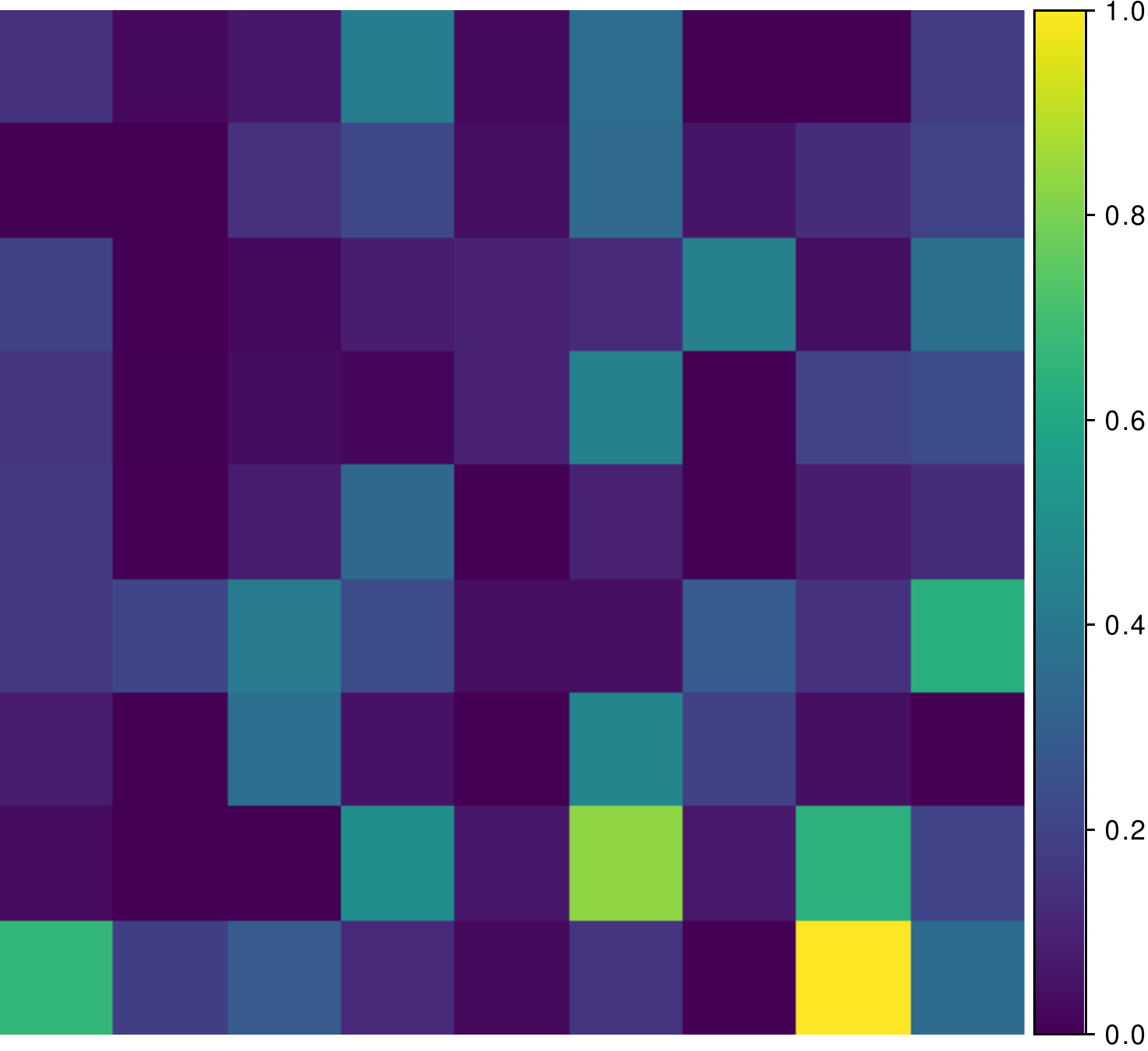}}}
\subfloat[]{{\includegraphics[scale=0.18]{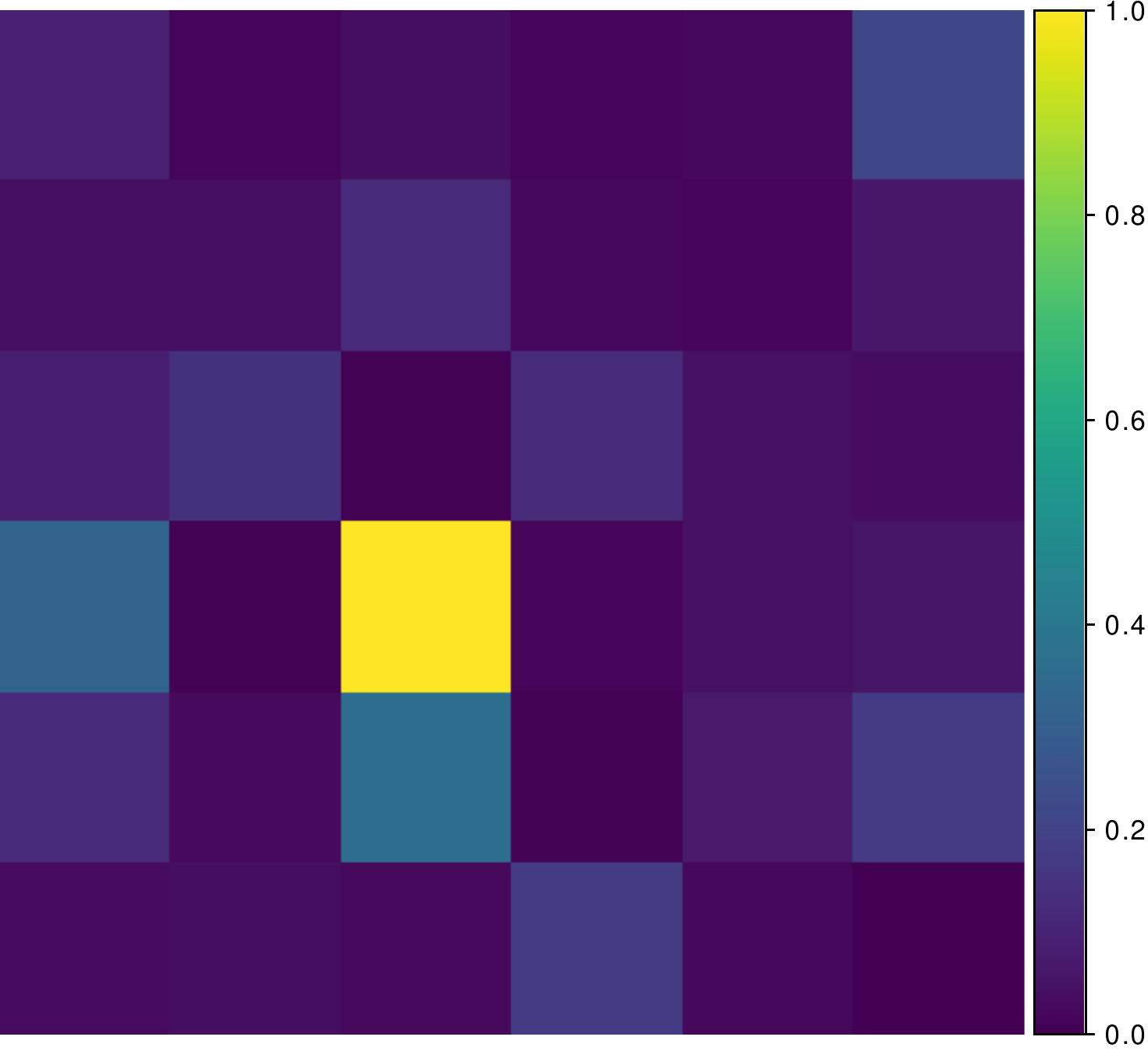}}}
\\\vskip 0.5pt plus 0.25fil
\subfloat[]{{\includegraphics[scale=0.18]{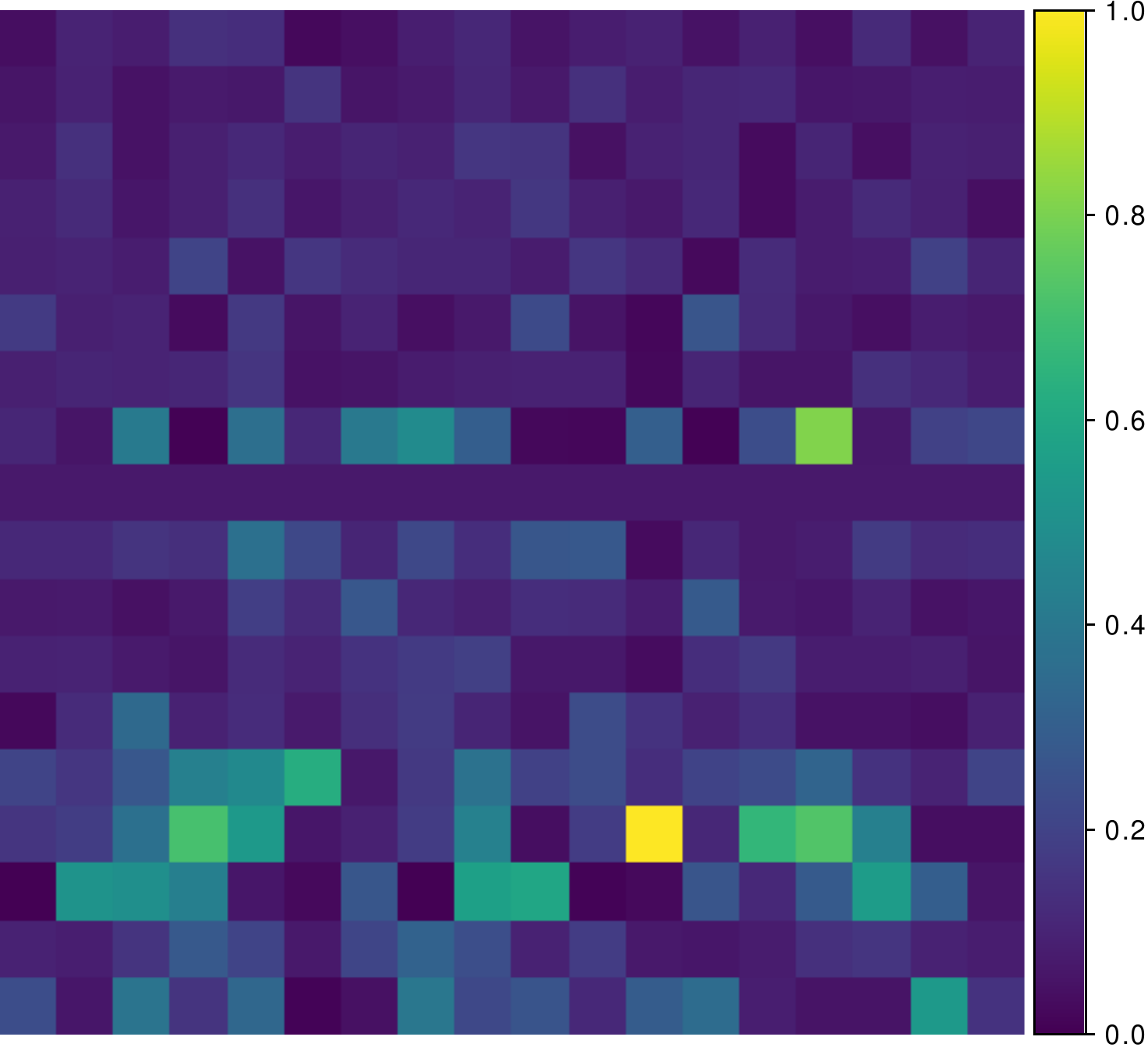}}}
\subfloat[]{{\includegraphics[scale=0.18]{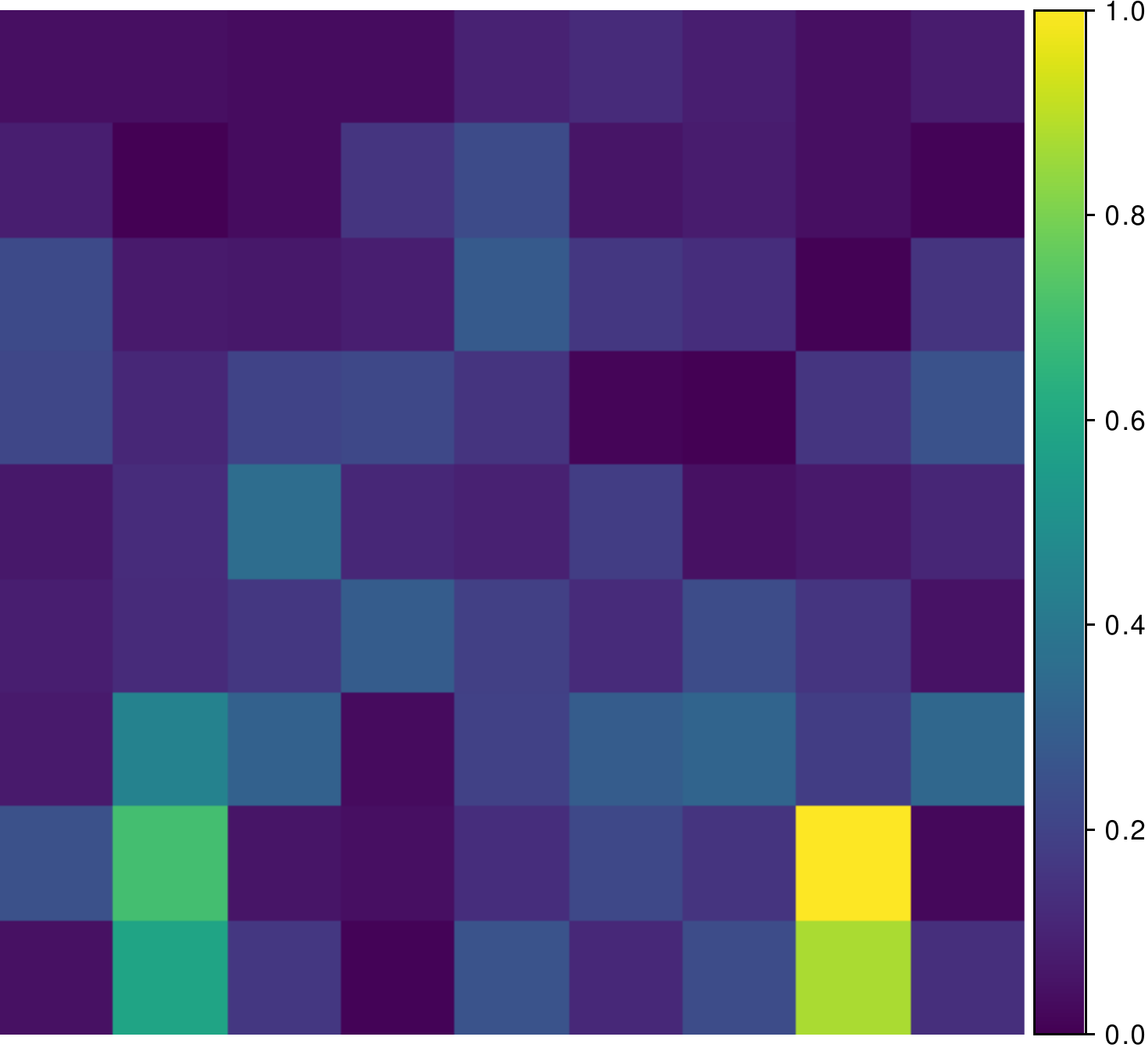}}}
\subfloat[]{{\includegraphics[scale=0.18]{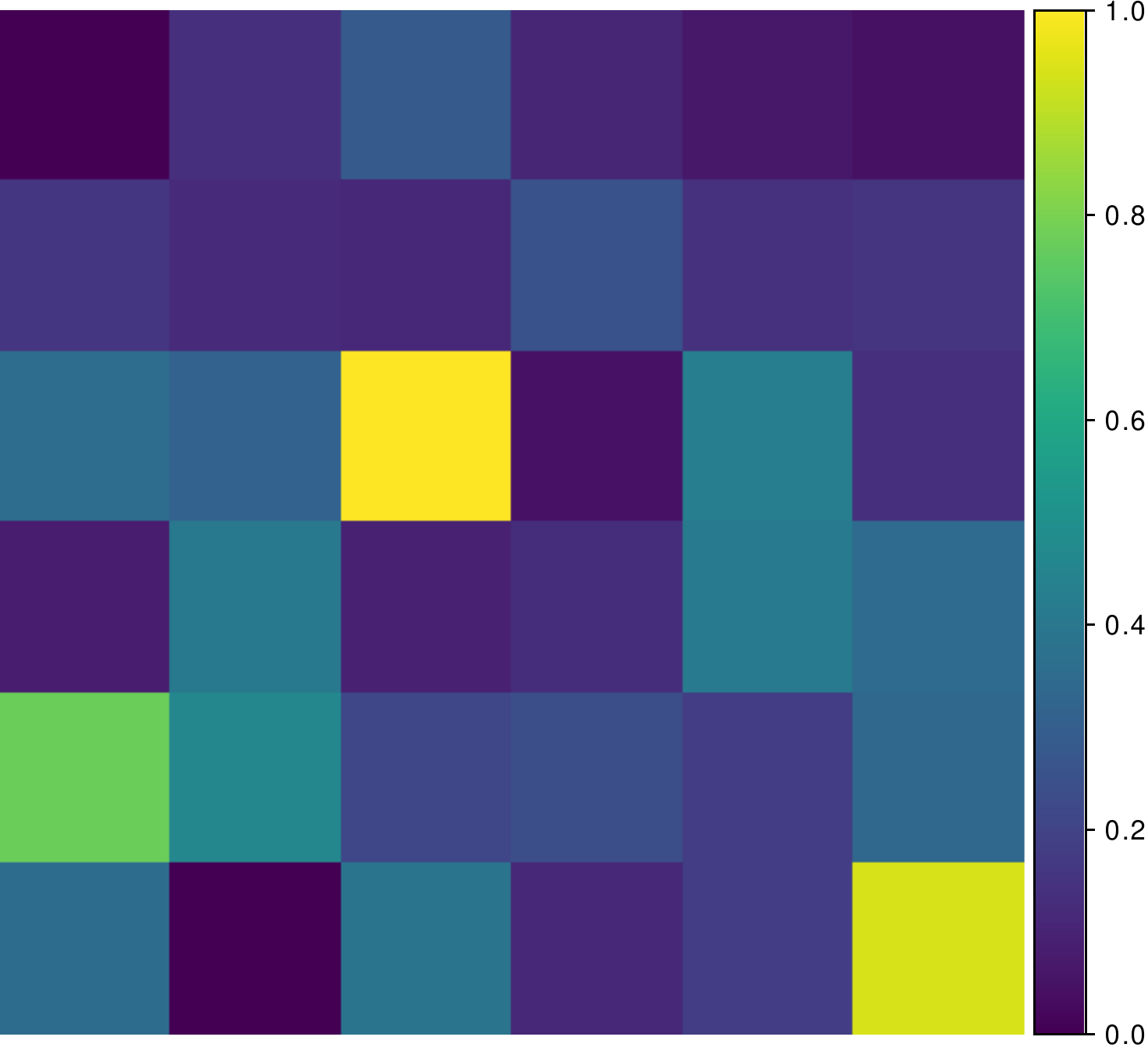}}}
\caption{Results of the spatial occlusion analysis using different mask sizes. (a,d): mask size 1$\times$1. (b,e): mask size 2$\times$2. (c,f): mask size of 3$\times$3. For both Att-Unistream and Att-Multistream models, the target city is Paris. The Att-Unistream and Att-Multistream models were trained to predict the temperature and wind speed, respectively.}
\label{fig:occlusion_squares}
\end{figure}

Apart from determining the important features or cities only, occlusion analysis can also be useful to determine whether a group of cities or features is important as well. We applied the same process described above, however instead of using vector masks, we used square patch matrices that are slided along both the rows and columns directions, without overlapping. Fig. \ref{fig:occlusion_squares} shows the visualization of this occlusion analysis, where the reference MSE corresponds to the error between the prediction and the actual data of the cities of Paris. The top and bottom rows show the visualization of this analysis for the Att-Unistream and Att-Multistream models, respectively. These models were trained for a 6 days ahead prediction of the temperature and wind speed. The first, second, and third columns use patch sizes of 1$\times$1, 2$\times$2, and 3$\times$3, respectively. Brighter colors correspond to features and cities that are more important. A first observation is that occlusions seem decisive about the important features and cities since only a specific mask region is brighter than the other regions in each occlusion map. The important features to predict the temperature shown in subfigures (a), (b), and (c) are the temperature and the pressure. These findings bring complementary information to the outcomes shown by subfigure (a) of Fig. \ref{fig:occ_analysis_m1}. Subfigures (d), (e), and (f) also reveal complementary information when compared to the analysis shown in Fig. \ref{fig:occ_analysis_m3}. Among the weather features that are important to predict the wind speed, we can find the maximum wind speed and the wind speed. Moreover, Brussels as well as cities near the sea like Barcelona or Amsterdam are critical cities for wind speed prediction. 

\begin{figure}[!t]
\centering
\subfloat[]{{\includegraphics[width=\linewidth]{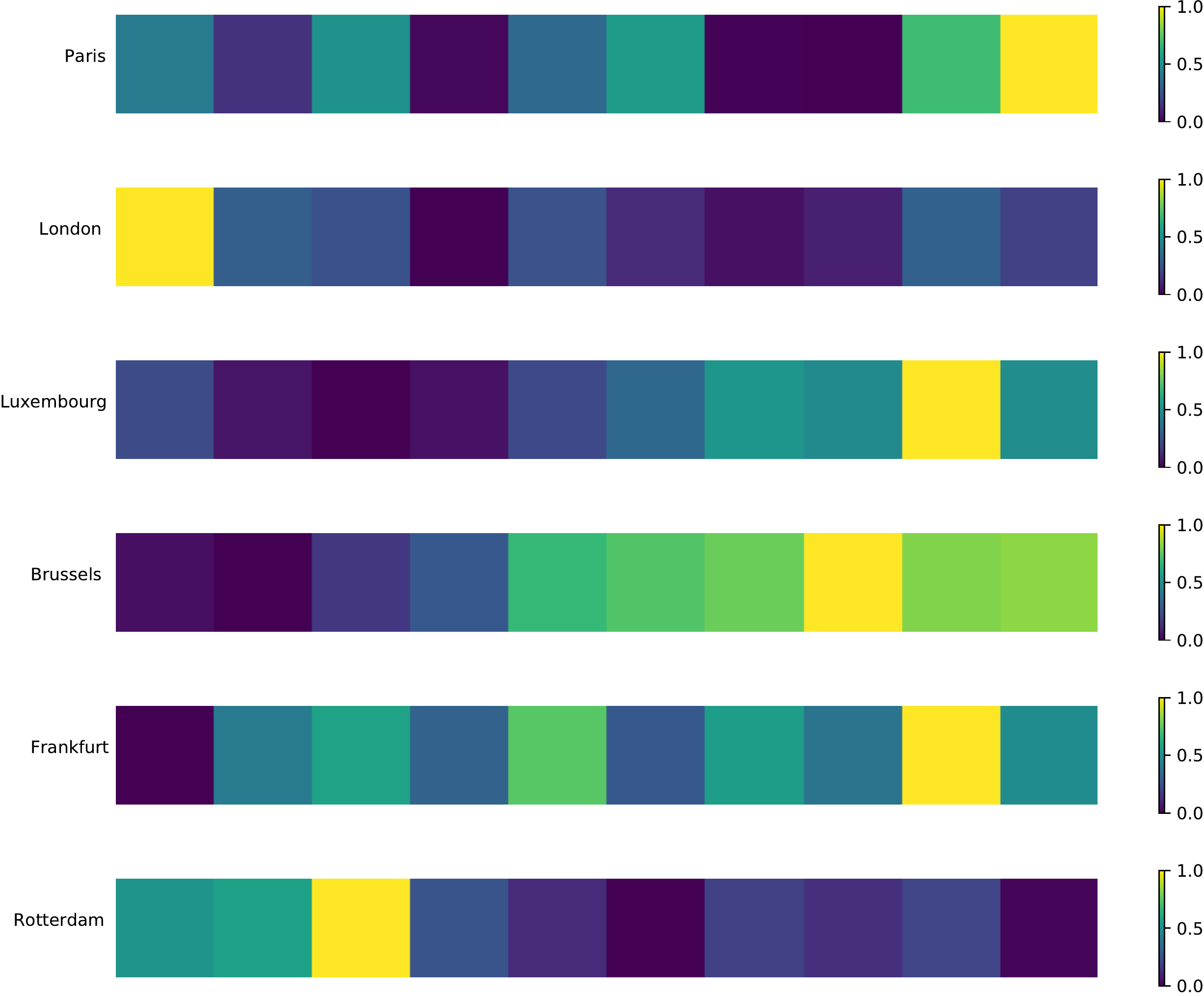}}}
\\\vskip 0.5pt plus 0.25fil
\subfloat[]{{\includegraphics[width=\linewidth]{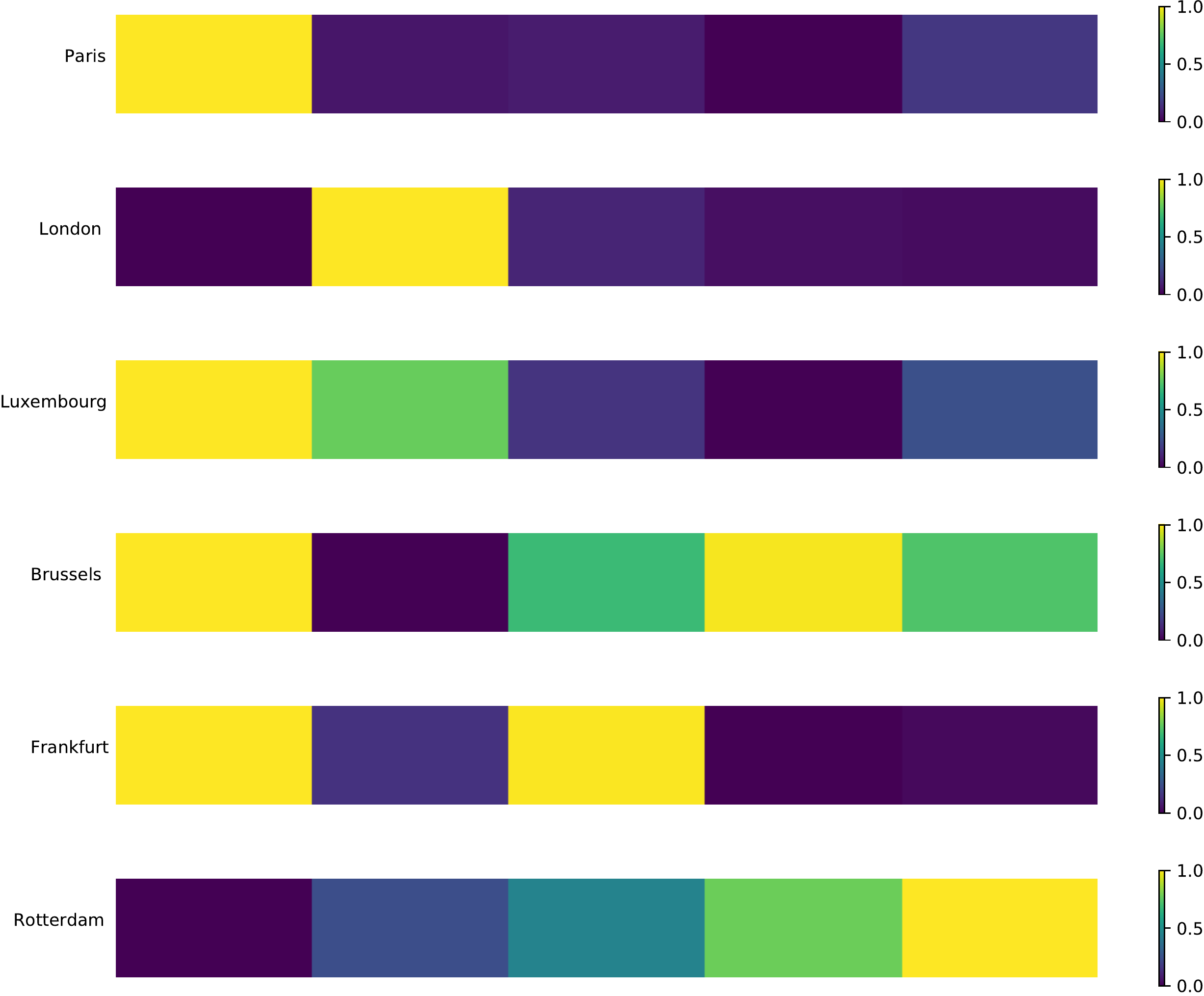}}}
\caption{Temporal occlusion analysis visualization of the Att-Unistream (a) and Att-Multistream (b) models. These models were trained to perform the 6 days ahead prediction of temperature and wind speed, respectively.}
\label{fig:temporal_occ_analysis}
\end{figure}

\begin{figure}[!t]
\centering
\subfloat[]{{\includegraphics[width=0.3\linewidth]{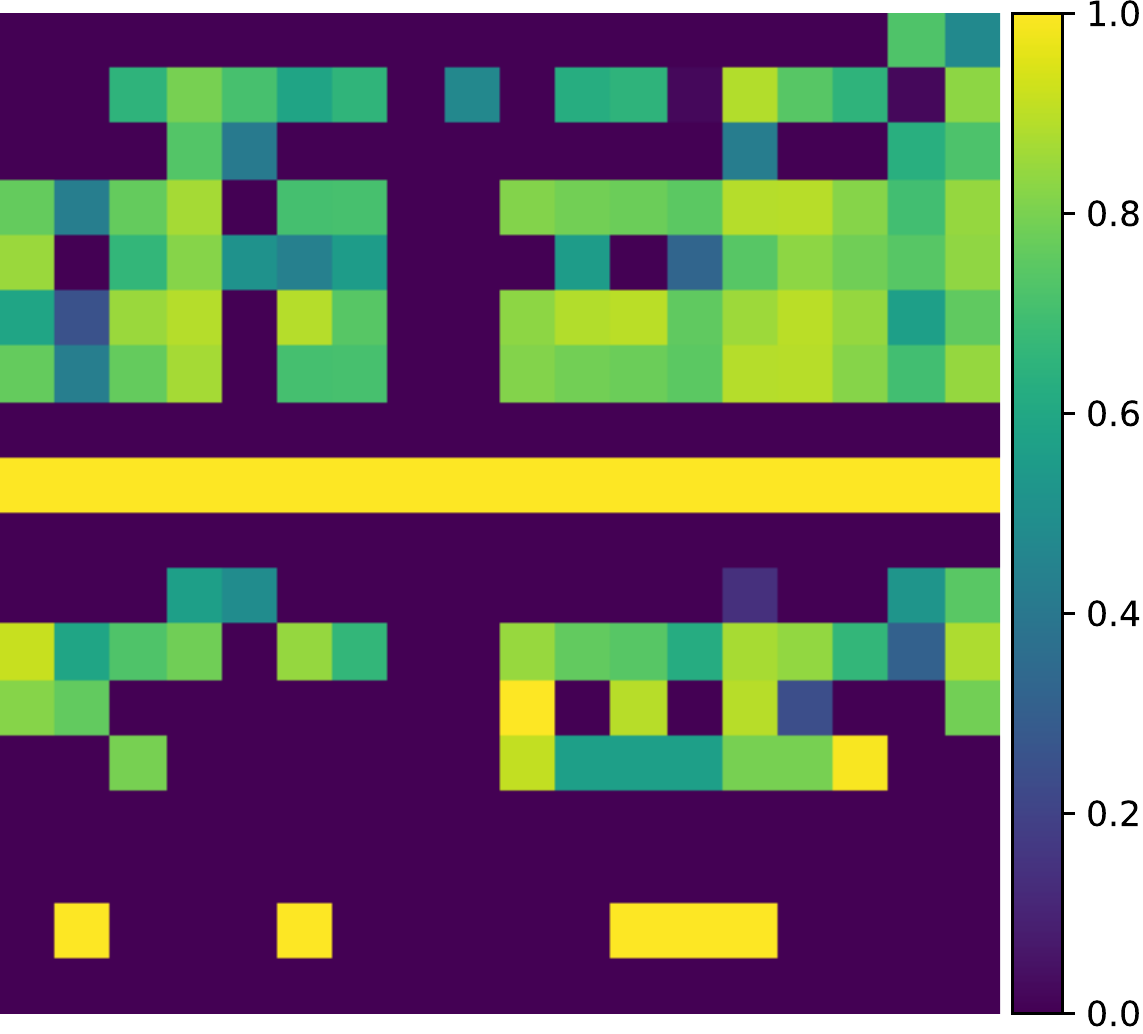}}}
\subfloat[]{{\includegraphics[width=0.3\linewidth]{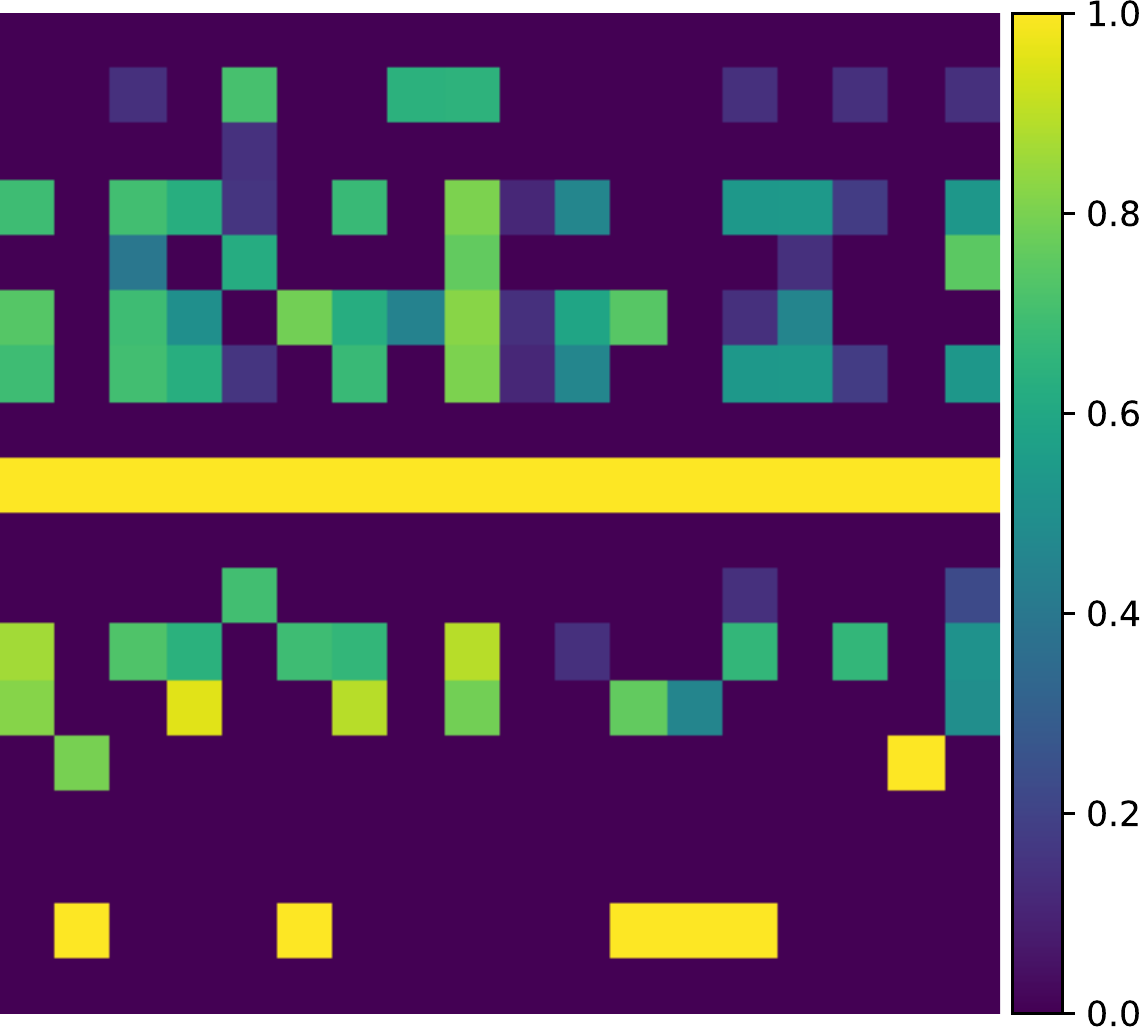}}}
\subfloat[]{{\includegraphics[width=0.3\linewidth]{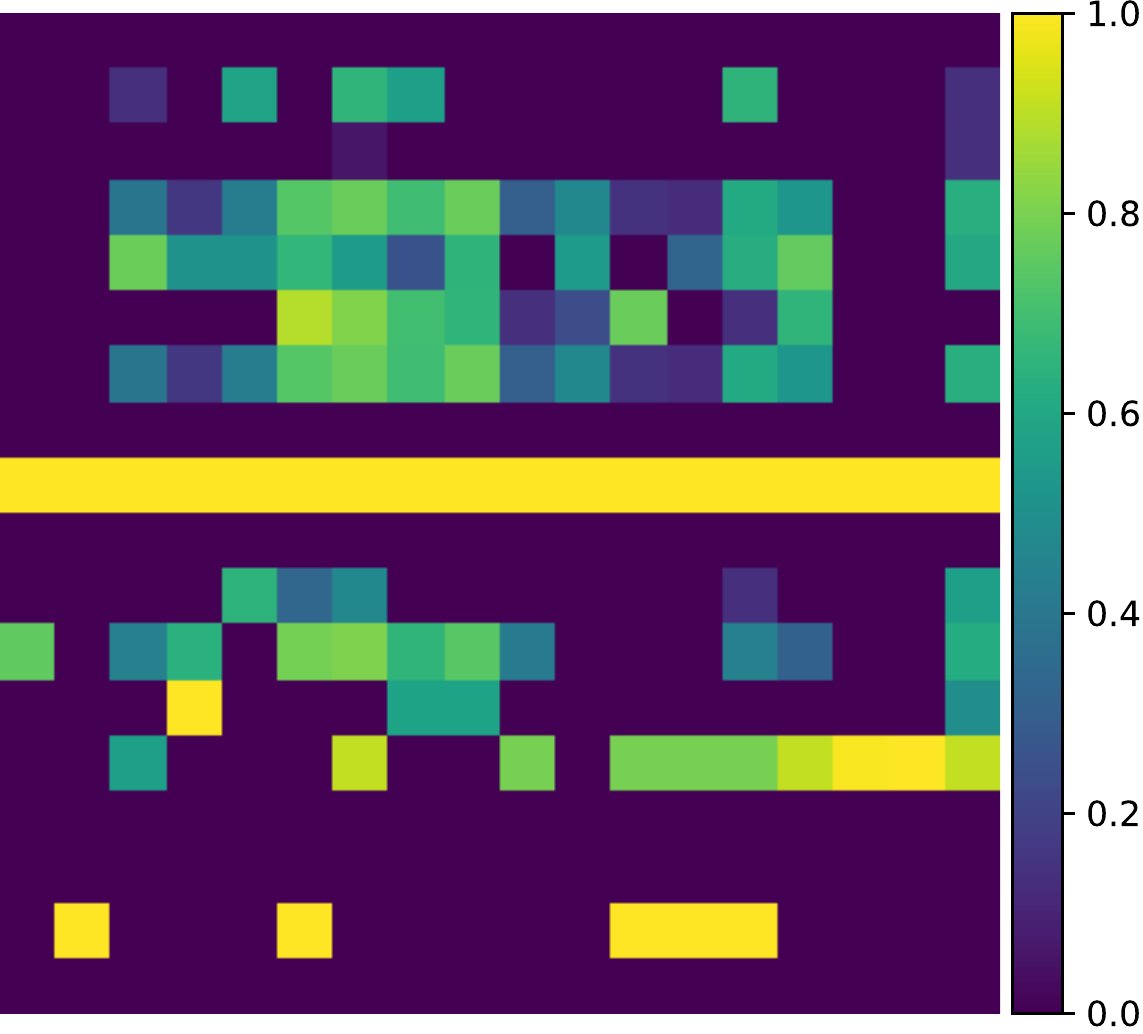}}}
\caption{Score maximization maps of the Att-Unistream for the 1\textsuperscript{st} (a), 5\textsuperscript{th} (b), and 10\textsuperscript{th} (c) lags. This model was trained to predict the wind speed.}
\label{fig:act_max_map_nsconv+lstm}
\end{figure}

Fig. \ref{fig:temporal_occ_analysis} displays the visualization of the temporal occlusion analysis where brighter values mean more relevant lags. As previously mentioned, this approach aims at finding which lag contributes the most to a minimal error between the actual and prediction data. The leftmost region corresponds to the oldest lags while the rightmost one are the most recent lags. For the Att-Unistream model, Fig. \ref{fig:temporal_occ_analysis} (a), there is an emphasis on the recent lags, with the exception of the cities of London and Rotterdam. However, for the Att-Multistream model in Fig. \ref{fig:temporal_occ_analysis} (b), the older lags yield consistently highest importance, with the exception of Rotterdam. Interestingly enough, the 1st and 2nd lags seem the most important ones for the cities of Paris, Luxembourg, Brussels, and Frankfurt.

\subsubsection{Score maximization}
Here we show the visualization of the score maximization introduced in section \ref{ssec:activ_max}. Fig. \ref{fig:act_max_map_nsconv+lstm} are the score maximization maps that show the relevant weather features for all the cities of the Att-Unistream model, trained to predict six days ahead daily average temperature. Higher pixel values in the score maps refer to a more important city-feature pair. There is no discernible pattern except the consistent feature across the three lags. This weather feature is the visibility. Moreover, the pressure feature seems important as well, but to a lesser extent.

\section{Conclusion}\label{sec:conclusion}
In this paper, two deep neural networks architectures have been proposed and investigated to perform weather elements forecasting. Moreover, a self-attention mechanism proved to be beneficial to these models since it consistently improved the results. From the analysis of the experimental results, it has been shown that a multistream input representation is globally more suitable for this task. In addition, interpretability techniques such as occlusion analysis and score maximization have been used to extract the most relevant input features (i.e. weather features and cities). These methods revealed that in general, Brussels is important for the prediction of temperature since it is located in the center of the target cities, while cities near the sea like Barcelona or Amsterdam are more suitable for wind speed prediction. It has also been shown that dew point is an important feature for the prediction of the temperature while the maximum wind speed and the condition heavily influence the wind speed prediction. From a temporal perspective, each model favors specific lags, with the unistream model having more emphasis on the recent lags and the multistream model favouring the older lags. The data and code used can be found at  \href{https://github.com/IsmailAlaouiAbdellaoui/weather-forecasting-explanable-recurrent-convolutional-NN}{github.com/IsmailAlaouiAbdellaoui/weather-forecasting-explanable-recurrent-convolutional-nn}.

\section*{Acknowledgment}
Simulations were performed with computing resources granted by RWTH Aachen University and Cloud TPUs from Google's TensorFlow Research Cloud (TFRC).

\bibliography{Main}

\end{document}